# Construction of extra-large scale screening tools for risks of severe mental illnesses using real world healthcare data

Dianbo Liu,PhD[1,7], Karmel W. Choi,PhD [2], Paulo Lizano,MD[3,4], William Yuan,PhD[1], Kun-Hsing Yu,PhD,MD[1], Jordan W. Smoller, MD, ScD[5,6], Isaac Kohane,PhD,MD[1,*]

1. Department of Biomedical Informatics, Harvard Medical School, Boston, MA,USA
2. Center for Human Genetics Research,Massachusetts General Hospital,Boston, MA, USA
3. Department of Psychiatry,Beth Israel Deaconess Medical Center, Boston, MA, USA
4. Department of Psychiatry, Harvard Medical School, Boston, MA, USA
5. Center for Precision Psychiatry, Massachusetts General Hospital, Boston, MA, USA
6. Psychiatric and Neurodevelopmental Genetics Unit, Massachusetts General Hospital, Boston, MA, USA
7. Broad Institute of MIT and Harvard, MA, USA

*Corresponds to: Isaac_Kohane@hms.harvard.edu

Word count: 6128 (including references)

## Key points

**Question:** Is it possible to build machine learning based tools to conduct extra-large scale population-level risk screening for severe mental illnesses (SMIs)?

**Finding:** Using data from beneficiaries from a nationwide commercial healthcare insurer with 77.4 million members and data from patients from EHRs from eight academic hospitals based in the U.S., a scalable machine learning based tool was developed to conduct population-level risk screening for SMIs, including schizophrenia, schizoaffective disorders, psychosis, and bipolar disorders. The obtained predictive models for SMIs achieved AUCROC of 0.76, specificity of 79.1% and sensitivity of 61.9% on an independent test set of an all-age case-control cohort from insurance claim data, and AUCROC of 0.83, specificity of 85.1% and sensitivity of 66.4% using EHR data.

**Meaning:** Performance of our SMI prediction models constructed using health insurance claims or EHR data suggest feasibility of using real world healthcare data for large scale screening of SMIs in the general population.

# Abstract

**Importance:** The prevalence of severe mental illnesses (SMIs) in the United States is approximately 3% of the whole population. The ability to conduct risk screening of SMIs at large scale could inform early prevention and treatment.

**Objective:** A scalable machine learning based tool was developed to conduct population-level risk screening for SMIs, including schizophrenia, schizoaffective disorders, psychosis, and bipolar disorders,using 1) healthcare insurance claims and 2) electronic health records (EHRs).

**Design, setting and participants:** Data from beneficiaries from a nationwide commercial healthcare insurer with 77.4 million members and data from patients from EHRs from eight academic hospitals based in the U.S. were used. First, the predictive models were constructed and tested using data in case-control cohorts from insurance claims or EHR data. Second, performance of the predictive models across data sources were analyzed. Third, as an illustrative application, the models were further trained to predict risks of SMIs among 18-year old young adults and individuals with substance associated conditions.

**Main outcomes and measures**: Machine learning-based predictive models for SMIs in the general population were built based on insurance claims and EHR.

**Results:** A total of 301,221 patients with SMIs and 2,439,890 control individuals were retrieved from the nationwide health insurance claim database in the U.S.. A total of 59,319 patients with SMIs and 297,993 control individuals were retrieved from EHRs spanning eight different hospitals from a major integrated healthcare system in Massachusetts, U.S.. The obtained predictive models for SMIs achieved AUCROC of 0.76, specificity of 79.1% and sensitivity of 61.9% on an independent test set of an all-age case-control cohort from insurance claim data, and AUCROC of 0.83, specificity of 85.1% and sensitivity of 66.4% using EHR data. The fine tuned models for specific use case scenarios outperformed two rule based benchmark methods when predicting 12-month risk of SMIs among 18-year old young adults but had inferior performance to benchmark methods when predicting SMIs among individuals with substance associated conditions in claims data.

**Conclusion:** Performance of our SMI prediction models constructed using health insurance claims or EHR data suggest feasibility of using real world healthcare data for large scale screening of SMIs in the general population. In addition, our analysis showed cross data source generalizability of machine learning models trained on real world healthcare data. Models constructed from insurance claims appear to be transferable to EHR cohorts and vice versa.

# Introduction

Approximately 3% of the U.S. population is affected by SMIs such as schizophrenia, schizoaffective disorders, other psychotic disorders, and bipolar disorder, which result not only in major social and occupational impairments but also in substantial societal costs(Hughes et al. 2016; Klinkenberg et al. 2003). Given that the effectiveness of available treatments for SMIs is enhanced by early identification and treatment engagement (Millan et al. 2016; McH and Slavney 2011; Roberts 2019; Jääskeläinen et al. 2013), there is considerable interest in identifying patients at high risk at the population level in order to target preventive care (Chen, Ventriglio, and Bhugra 2019; Riecher-Rössler and McGorry 2016; Cannon et al. 2016). In addition, there is evidence that early detection and treatment of SMIs can improve patient outcomes(Larsen et al. 2011; Díaz-Caneja et al. 2015; Marshall et al. 2005; Perkins et al. 2005; Hegelstad et al. 2012; Larsen et al. 2005; Melle et al. 2004). Therefore, construction of large-scale population level screening tools for risks of SMIs could be of substantial value for population health management.

Strategies to identify risk for schizophrenia, psychosis and other psychiatric disorders have largely relied on in-person clinical assessments.(A. R. Yung et al. 1996; Fusar-Poli et al. 2015; A. Yung and O'Dwyer 2006; Krebs et al. 2014; French and Morrison 2004). Early treatment interventions based on these strategies have demonstrated reduction of future psychotic disorders((Stafford et al. 2013; Häfner and Maurer 2001; Cannon et al. 2016; Fusar-Poli et al. 2017). However, these strategies usually require either in-person interviews or questionnaires, and focus on prodromal psychological disorders or patients already identified as clinically at risk, limiting their scalability and generalizability to a broader population(Cannon et al. 2016). In recent years, electronic health record (EHR) data and healthcare insurance claims data have become widely available and have begun to be used in monitoring risks of mental illness(Ruetsch, Un, and Waters 2018; Smoller 2017). The value of EHR and insurance claims data for psychiatric phenotyping and risk prediction lie in the breadth and depth of the information they contain and their potential utility for large-scale, low cost screening for risk of psychiatric outcomes (Mandel et al. 2016; Smoller 2017).

Despite the potential opportunities afforded by EHR and insurance claims databases for psychiatric applications, several challenges remain to be addressed.. First of all, due to operational and regulatory reasons, EHR and claims data are often siloed. For example, few studies have used insurance claim data and EHR data from multiple institutions or organizations to validate the power and potential of using each of these data sources for SMI prediction at a national scale. Though EHR and insurance claim data share many common variables, they are fundamentally different in the purpose of data collection and distribution. Insurance claims data are primarily used for billing purposes and capture clinical information across many healthcare providers. In comparison, EHR data, while also providing billing information, are primarily used for facilitating clinical practice in hospitals and often reflect information about patients from a single or a small number of healthcare providers. Secondly, though EHR and claims data include a larger variety of information compared with interviews and questionnaires, they usually contain less depth of phenotyping for a disease of interest. Lastly, because EHR and claims data from different institutions often use different ontologies, granularity, and data structures, it may be challenging to apply methods developed in one institution in another(Mandl and Kohane 2017; Kohane and Altman 2005; Liu, Dligach, and Miller 2019; Shao et al., n.d.).

This study investigates, for the first time to our knowledge, the potential of utilizing national scale insurance claims data and EHR data from multiple institutions for large scale risk prediction of SMIs. Our SMI risk prediction models use demographic information, medication and prior diagnoses of individuals to predict future onset of SMI In the model construction process, predictive models were trained using a two-step cross-data-source training strategy using case-control cohorts retrieved from EHR and insurance claims databases. The performance of the predictive models obtained were also checked in two steps. First, predictive model performance was validated in out-of-sample data from the case control cohorts. Second, we explored the robustness and generalizability of each model by applying models trained on insurance claims onto EHR data and vice versa. In addition, the models were fine-tuned and their performance was verified in two specific illustrative use cases. The first use case was predicting risks of SMI among 18-year-old young adults, an age at which incidence rates of many psychiatric disorders are high and a larger portion of patients are changing healthcare providers and health insurance providers(Mechanic 1979; Campbell 2017; Test et al. 1985). The second illustrative use case was predicting the risks of SMI among individuals with substance use disorder (Moschetti et al. 2015; Drake and Wallach 1989).

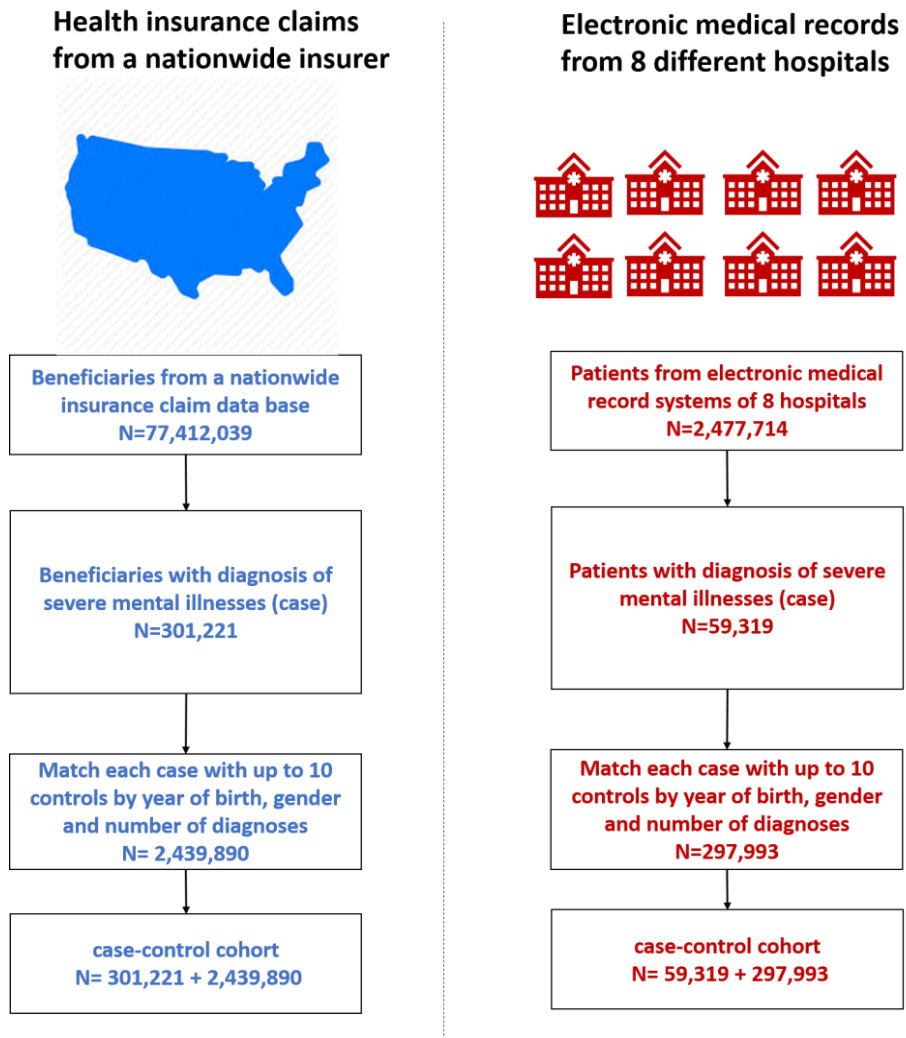

**Figure 1.** Two data sources were used in this study: insurance claims data from a nationwide healthcare insurer in the U.S., and electronic medical records from 8 different hospitals located in  estern Massachusetts, U.S.. The case-control cohorts included all patients with SMIs diagnoses and control patients matched by age, gender, and number of diagnoses before starting the observational period ( See method section for details).

# Methods

**Study subjects and clinical characteristics.** Data from two sources were included in this study. The first data source is health insurance claim data from a private healthcare insurer, Aetna, with medical information on more than 77 million insurance beneficiaries in total from across the United States. The Aetna database contains health-related insurance claims from 2008-01-01 to 2019-07-01. The average enrollment period was 25 months (SD 27.1) In the case-control study, all patients with SMIs diagnoses were retrieved to define cases ( see next section for details). A detailed list of SMIs diagnoses used in this study for case definitions is included in the next section. Demographic information (year of birth and gender), diagnosis codes (ICD-9 or ICD-10) and medication fills (National Drug Code or NDC) in a 12-month observational period were included as input features for the predictive model. A gap period with random length between 14 and 365 days was introduced between the ends of observation periods and the first onset date of SMIs to avoid potential statistical biases and enable the trained model to make predictions on patients at different stages of disease development (Figure 2)(Yuan et al. 2021). Up to 10 control individuals without severe mental illnesses were matched to each patient with SMIs by age, gender, and number of diagnoses before starting the observational periods of corresponding cases. Observational periods of matched control individuals were assigned as the same period as corresponding matched SMIs cases.

The second data source is EHR data from eight hospitals in the Mass General Brigham (MGB) healthcare system based in Massachusetts, USA. Data were extracted from the MGB Research Patient Data Registry (RPDR). A total of 2,477,714 patients' electronic medical records were available. The cases and control were defined in the same way as in the claims data. Patients with at least one diagnosis of any disease during the 12-month observational period were included in this study. Same as in the claims data, demographic information, diagnosis codes and medication in the observational period and used as input features. It is worth pointing out that these input features were used in this study because they are readily available in both datasource and overlaps with predictors of SMIs(Cannon et al. 2016; American Psychiatric Association 2013; Glass 2009).

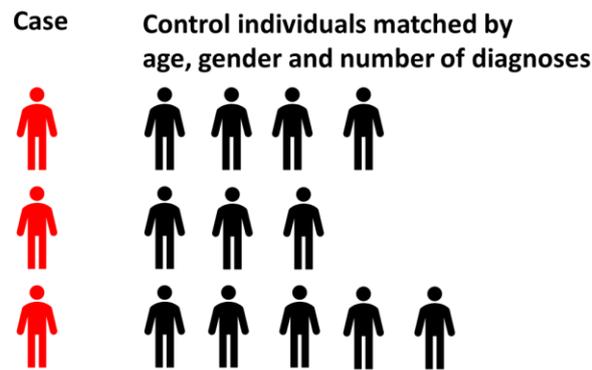

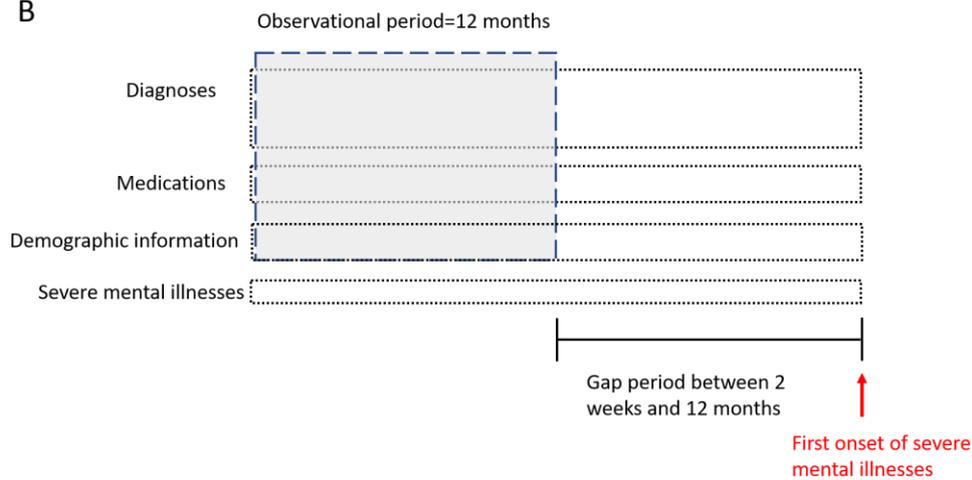

**Figure 2. Case control cohort from either insurance claims data or electronic records data.** Cases comprised patients with one or more ICD codes for SMIs (see Materials and method for details). Demographic information, diagnosis, and medication in a 12-month observational time window before the first recorded onset of SMIs were used as predictive model inputs. A gap period of random length between 0 and 12 months was included to introduce heterogeneity. Each case of SMIs was matched with up to 10 control individuals with no SMIs diagnoses matched by year of birth, gender, and number of any diagnoses before the observational periods.

**Case definition.** Diagnoses in both the claims and EHR datasets were originally provided as International Classification of Diseases, Ninth Revision and Tenth Revision (ICD-9 and ICD-10) codes. ICD9/10 codes for SMIs were selected according to the mapping between ICD 9/10 codes and Phecodes from PheWASCatalog (https://phewascatalog.org/). Phecodes represent a hierarchical grouping of ICD codes into a reduced set of clinically similar codes(Bastarache et al. 2018). An individual was defined as having a disorder of interest (SMIs) if he or she had at least one Phecode for the disorder. If an individual had multiple diagnoses for SMIs, the date of the first occurrence was used to define gap and observational periods. Schizophrenia (Phecode 295.1), Schizoaffective disorder (295.1), psychosis (295.3) and bipolar disorders (296.1) were considered as SMIs in this study.

**Specific use cases.** When predicting risks of SMIs among 18-year-old young adults, cohorts of individuals with records in the 24-month period around their 18th birthdays from EHR or insurance claims database were used. Individuals in these cohorts were excluded from the all-age-case-control cohorts described in the sections above. Demographics, diagnoses and medications from the 12-month observational period before the 18th birthday of the individual were used as input features into the model to predict presence of SMIs diagnoses in a 12-month test period after their 18th birthdays.

When predicting risks of SMIs among individuals with existing diagnoses of substance related conditions, demographics, diagnoses and medications from a 12-month observational period before (and including) the date of the first occurrence of diagnosis codes of substance related conditions was used as input features to predict presence of SMIs diagnoses in a 12-month test period after it. Substance related conditions in this analysis included substance addiction and disorder (Phecode 316), alcoholism and alcohol related disorders (Phecode 317), and tobacco use disorders (Phecode 318).

**Data access and IRB.** Data access to Aetna data warehouse has been granted by the Aetna data warehouse at Harvard Medical School. Access to EHR from the Mass General Brigham (MGB) healthcare system was approved by the MGB IRB.

**Predictive model development, performance measurement, and benchmark methods.** Models were trained and tested separately for the claims and EHR dataset. In each dataset, individuals were randomly split into a training set (60%), validation set (10%), and test set (30%). Predictive models were trained using the training set, and hyperparameters, such as the number of training epochs, were adjusted according to performance on the validation set. The performance was measured by AUCROC (area under the receiver operating characteristic curve, range between 0 and 1, the closer to 1 the better the performance is), specificity (range between 0 and 1, the closer to 1 the better the performance is) and sensitivity (range between 0 and 1, the closer to 1 the better the performance is). When conducting cross dataset performance testing (ie, making predictions on EHR dataset using a model trained on only claims data and vice versa), only input features overlapped between the two data sources were used as model inputs.

A neural network based machine learning was used in this study for its ability to learn representation of complex patterns from data and flexibility in model design. Artificial neural networks are a class of machine learning models inspired by the biological neural networks and are able to learn hierarchical representation of complex patterns from data given enough sample sizes. The model structure consists of an embedding layer, an averaging layer, two feedforward layers and a single-unit output layer (Figure 3). Sigmoid activation function was used at the output layer and ReLu activation was used in hidden layers. A embedding converts each index input in the form of an integer ( eg. [16]) into a real value vector of fixed length (eg. [0.15,1.37,....,0.76]). The embedding layer of embedding size 300 was used for diagnosis and medication inputs because the inputs were very sparse for most individuals. The embedding layer together with the averaging layer significantly reduced the dimension of inputs into the feedforward layers. The embedding layer was trained together with other parts of the model using the same data. Therefore, no additional pre-training of embedding was conducted in this study. The Demographic information was input into the first feedforward layer directly. A threshold was set to maximize the Youden index for the binary classification task.

Two rule-based benchmark methods were included in this study to compare with machine learning-based predictive models. In the first benchmark method (benchmark 1), individuals with any diagnosis codes in the psychological condition category (PheCode 295 to 307), excluding SMIs, in the 12-month observational periods were predicted as having high risks of SMIs (positive in the binary classification) in the test period. In the second benchmark method (benchmark 2), individuals with any axis I diagnosis codes as defined in DSM IV, excluding SMIs, in the 12-month observational periods were predicted as having high risks of SMIs (positive) in the test period(Glass 2009; American Psychiatric Association 2013). Axis I diagnoses in observational period were used a benchmark predictive method because previous studies suggested that Axis I psychopathology were associated with elevated risk of SMIs (Shah et al. 2019; Addington et al. 2017), The list of of axis I diagnoses used in for benchmark 2 were adapted from Shah *et al.* 2019 and Addington *et al.* 2017. A detailed list of Axis I diagnoses and their Phecodes can be found in supplementary table 1.

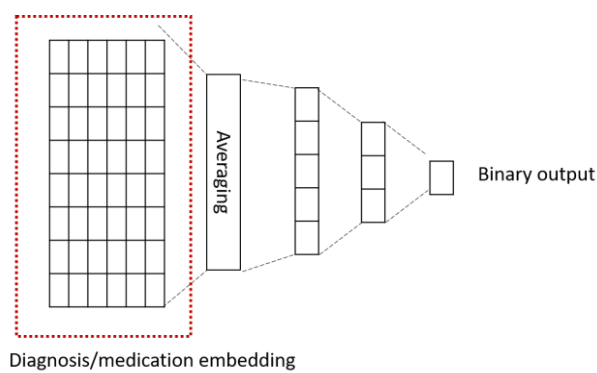

**Figure 3 Machine learning model.** Artificial neural networks are a class of machine learning models inspired by the biological neural networks and are able to learn hierarchical

representation of complex patterns from data given enough sample sizes. The artificial neural network we used in this study consisting of an embedding module and a feedforward module. The embedding module was included to deal with sparsity of diagnosis and medication code in the datasets (See method section for details).

## Data availability

Data access to Aetna data warehouse at Harvard Medical School and access to EHR from the Mass General Brigham (MGB) healthcare system are both restricted by regulations and laws and are, therefore, not publicly available. Python and R scripts used for the analysis will be available on Github upon publication of the article.

## Results

**Study Cohort.** In the claims dataset, a total of 301,221 patients with SMIs and 2,439,890 control individuals without SMIs matched by age, gender, and number of diagnoses were retrieved. In the EHR dataset, a total of 59,319 patients with SMIs and 297,993 control individuals were retrieved (Figure 3).

**Accuracy of SMIs risk screening model constructed using claims data.** An artificial neural network based SMIs risk screening model was trained using the case-control cohort obtained from the claims database. The model takes diagnosis, medication from the observational period and demographic information as inputs and estimates the risks of SMIs. Throughout this study, training a machine learning model refers to the process to adjust model parameters so that the predicted values by the model become closer to the ground truth labels provided in the data. Performance of the SMIs risk screening model was tested on randomly selected out-of-sample data from the same cohort. We also included performance of two rule-based benchmark methods for comparison (see details in Materials and Methods).

The artificial neural network SMIs risk screening model achieved AUCROC of 0.76, specificity of 79.1% and sensitivity of 61.9%, on the test set. In comparison, benchmark method 1 achieved specificity of 80.4% and sensitivity of 40.8%. Benchmark method 2 achieved specificity of 78.0% and sensitivity of 40.0% (AUCROC not available for both benchmark methods) (Table 1).

**Accuracy of SMIs screening model using electronic medical record.** When using the case-control cohort from the EHR database for training and testing, the artificial neural network based SMIs risk screening model achieved AUCROC of 0.83, specificity of 85.1% and sensitivity of 66.4% on the test set. In comparison, benchmark method 1 achieved specificity of 79.4% and sensitivity of 45.0% and benchmark method 2 achieved specificity of 79.4% and sensitivity of 37.6%. Our method outperformed both benchmark methods when using EHR data (Table 1).

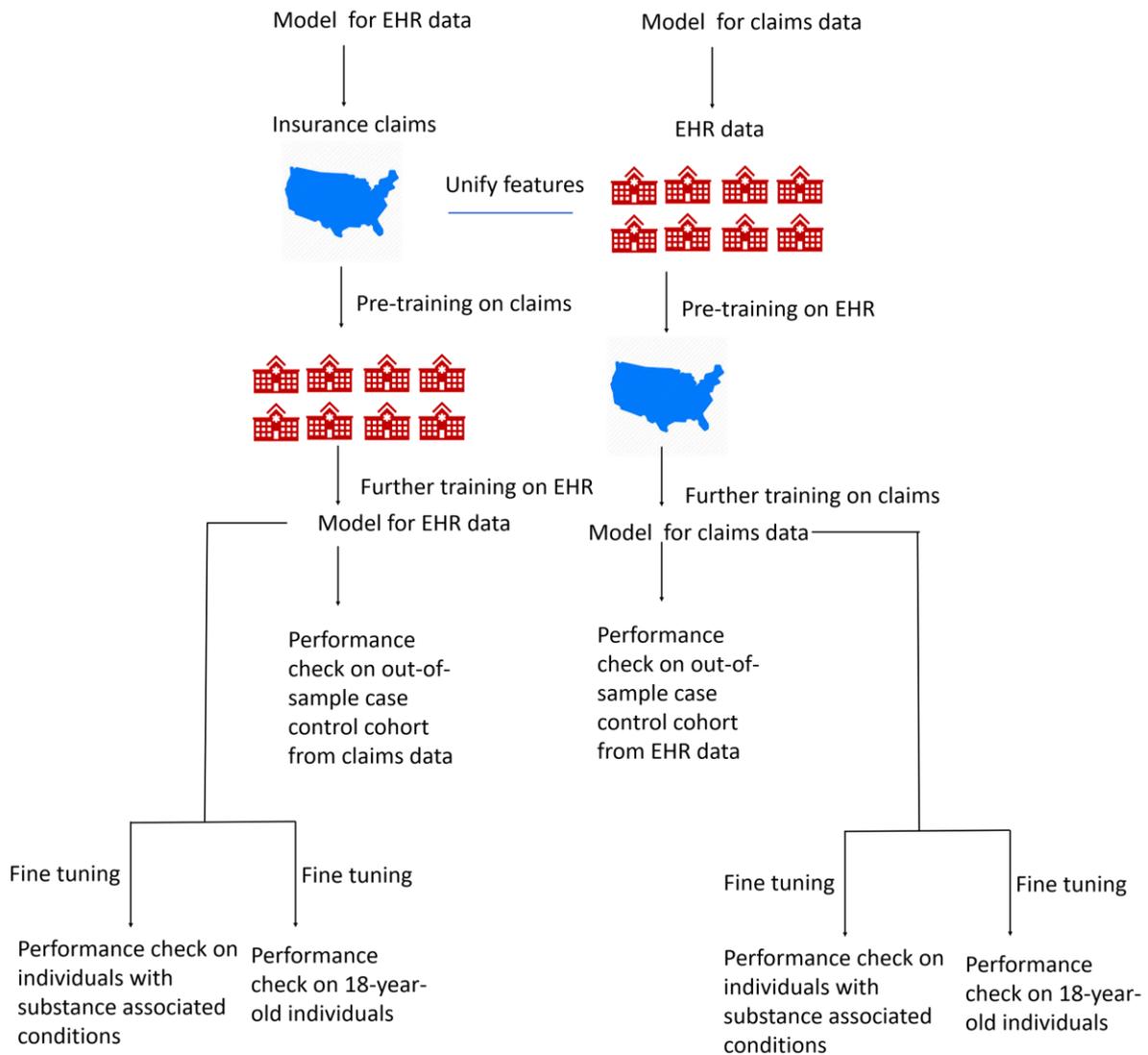

**Figure 3.** Training and performance testing of SMIs risk screening machine learning models.

**Cross-data-source performance of SMIs screening models.** To explore if a model trained on claims data could be applied to EHRs and vice versa, we conducted two sets of experiments. When the model was trained on claims data and tested on EHR, it achieved AUCROC of 0.75, specificity of 79.0% and sensitivity of 62.2%. On the other hand, when the model was trained on EHR and tested on claims data, it achieved AUCROC of 0.64, specificity of 78.4% and sensitivity of 44.9% (Table 2).

Though, the cross-data-source performance of SMIs screening models are inferior to within-data-source experiments, the results suggest claims data and EHR share some similarity between each other for SMIs. The similarity between these two data sources can be utilized to improve screening models. Based on this observation, we tried to further improve the screening using a two-step model training strategy. First, the model was pre-trained using EHR data to get a good model parameter initialization. Second, the obtained model was further trained with claims data. The model obtained is used for SMIs risk screening in claims data. The screening model for EHR data was constructed following the same logic. Our experimental results show this two-step training strategy does improve performance of both models, though only by a small magnitude (Table 3). In all subsequent sections, the models used were obtained using this two-step strategy.

**Table 1** performance of SMIs risk screening models trained using claims or EHR data

| Method | Data set | AUCROC | Sensitivity | Specificity | Prevalence |
|---|---|---|---|---|---|
| **Machine learning model** | **EHR** | **0.83** | **66.4%** | **85.1%** | **16.6%** |
| Benchmark 1 | EHR | NA | 45.0% | 79.4% | 16.6% |
| Benchmark 2 | EHR | NA | 37.6% | 79.4% | 16.6% |
| **Machine learning model** | **claims** | **0.76** | **61.9%** | **79.1%** | **10.9%** |
| Benchmark 1 | claims | NA | 40.8% | 80.4% | 10.9% |
| Benchmark 2 | claims | NA | 40.0% | 78.0% | 10.9% |

**Accuracy of SIMs risk screening models in specific use cases.** In the previous sections, models were trained and tested using case-control cohorts to maximize the number of data points that can be used for training. In order to have a better understanding of how the SMIs risk screening models we developed can be in public health and clinical settings, we tested their performances in two specific use case scenarios: 1) predicting risk of SMIs among young adults of age 18, as it is when incidence rates of many psychological disorders are high and a larger portion of patients are changing healthcare and health insurance providers (Mechanic 1979; Campbell 2017; Test et al. 1985), and 2) risk of SMIs among individuals who had substances associated conditions, as substance related mental illnesses is a severe problem. In each of these use case scenarios, the SMIs risk screening models trained using the method mentioned in the previous section was fine tuned using a training data set from a cohort specific to each use case and tested in a randomly selected out-of-sample test set from the same cohort.

When using claims data, the model screening SMIs risk among 18-year-old young adults achieved AUCROC of 0.71, specificity of 80.1% and sensitivity of 57.0%. The model screening SMIs risk among individuals with substance abuse achieved AUCROC of 0.53, specificity of 72.5% and sensitivity of 63.3%. When using EHR data, the model screening SMIs risk among 18-year-old young adults achieved AUCROC of 0.68, specificity of 89.0% and sensitivity of 37.4%. The model screening SMIs risk among individuals with substance abuse achieved AUCROC of 0.63, specificity of 59.4% and sensitivity of 72.7% , which is superior to baselines. (Table 4).

**Table 2** Cross-data-source performances of SMIs risk screening models

| Model | Test data | AUCROC | Sensitivity | Specificity | Prevalence in test set |
|---|---|---|---|---|---|
| Machine learning model trained using claims | EHR | 0.75 | 62.2% | 79.0% | 16.60% |
| Machine learning model trained using EHR | Claims | 0.64 | 44.9% | 78.4% | 10.90% |

**Table 3** performances of SMIs risk screening models when trained using our two-step training strategy

| Method | Data set | AUCROC | Sensitivity | Specificity | Prevalence |
|---|---|---|---|---|---|
| Machine learning model | EHR | 0.86 | 67.0% | 84.0% | 16.6% |
| Machine learning model | claims | 0.77 | 62.1% | 79.1% | 10.9% |

Table 4  performances of SMIs risk screening models in specific use cases

| Method | dataset/use case | AUCROC | Sensitivity | Specificity | Prevalence |
|---|---|---|---|---|---|
| **Machine learning model** | **EHR/18-year old** | **0.68** | **37.4%** | **89.0%** | **0.86%** |
| Benchmark 1 | EHR/18-year old | NA | 30.5% | 80.7% | 0.86% |
| Benchmark 2 | EHR/18-year old | NA | 29.3% | 79.8% | 0.86% |
| **Machine learning model** | **EHR/substances associated** | **0.63** | **72.7%** | **59.4%** | **3.3%** |
| Benchmark 1 | EHR/substances associated | NA | 54.5% | 48.3% | 3.3% |
| Benchmark 2 | EHR/substances associated | NA | 63.6% | 50.3% | 3.3% |
| **Machine learning model** | **Claims/18-year old** | **0.71** | **57.0%** | **80.1%** | **0.67%** |
| Benchmark 1 | Claims/18-year old | NA | 40.4% | 81.0% | 0.67% |
| Benchmark 2 | Claims/18-year old | NA | 44.5% | 80.7% | 0.67% |
| **Machine learning model** | **Claims/substances associated** | **0.53** | **63.3%** | **72.5%** | **4.5%** |
| Benchmark | Claims/sub | NA | 61.8% | 53.2% | 4.5% |
| 1* | stances associated | | | | |
| Benchmark 2* | Claims/substances associated | NA | 62.9% | 48.5% | 4.5% |

*Substance associated conditions (Phecode 316, 317,318) were excluded from criteria of benchmark 1 and 2 when applying to use case scenario 2.

## Discussion

In this study, we developed a machine learning based method to conduct large scale risk screening for SMIs using real world healthcare data. As insurance claims and EHRs are widely available in digitized forms in the vast majority of healthcare systems in the U.S. and many other countries, construction, implementation, and application of these screen tools requires less efforts and modification of existing systems compared with many existing interview base approaches(Cannon et al. 2016).

The large scale screening constructed using methods described in this study can be used for many public health and clinical purposes. For example, most insurance claims and EHR datasets include zip codes or county of the individuals. Our screening tool can be used to provide high resolution geographic hot spots for SMIs risks. This would allow healthcare providers or the government to design an implementation project to target those specific hotspots and to enhance the care of people with SMIs. Insurance claims and EHR data often come with information about gender, race, employer types etc. and the records in these systems are usually updated in a semi-real time manner. Using our method, it is possible to identify groups of individuals who urgently need care of SMIs in a semi-real time manner. This will help decision making in distribution and allocation of limited healthcare resources.

Both Diagnostic and Statistical Manual (DSM) IV and V did not provide a list of prodromal features of SMIs (George et al. 2017). We designed a benchmark SMIs risk prediction method using diagnosis of psychological condition and another benchmark using axis I diagnoses from DSM IV. Our machine learning based methods outperformed rule-based benchmarks when testing on holdout samples and predicting risks of SMIs among 18-year old young adults in both insurance claims and EHR data. These results suggested utilizing real world healthcare data to build large scale screening models for SMIs risks in the general population is a promising direction. However, the relatively low accuracy in both specific use case scenarios indicates that further improvement will be needed to build reliable and robust screening tools. One of such improvements could be including more input features such as clinical notes from healthcare providers in the model training to improve predictive power. Neurocognitive and functioning measures are often used in interview based SMIs screening but sometimes not available in insurance claims and EHR data ((Cannon et al. 2016). Inclusion of additional mental health related data into existing insurance and EHR databases will likely help improve screening models as well.  In addition, the performance of our model in predicting risks of SMIs among patients with existing substance associated conditions was inferior to both benchmarks in claims data. This observation suggested two things: 1) more data might be needed to train a model to predict SMIs among individuals with substance associated conditions, and 2) machine learning model trained on one use case ( eg. screening the general population) might not be directly transferable to another very different use case ( eg. individual with substance associated conditions). Therefore, use case-specific data and models will  be needed for public health and clinical practices.

In addition, we compare inter data source performance of the machine models. Performance of the model trained using insurance claims data was tested on  EHR data and vice versa. Our experimental results suggested that a model trained on EHR data  can be used on insurance claims data and achieved reasonable accuracy. The similar results were observed the other way around. Our finding suggested feasibility of using machine learning across different data sources. However, the reduced model performance in inter data source experiment compared with intra data source experiment still indicate the fundamental differences in these two data sources, though the variables in the two sources are similar.

In summary, using a nationwide medical claim dataset and EHR data from 8 different hospitals, we were able to develop scalable risk screening tool of SMIs to address the challenges of using multiple existing digitized real world healthcare datasets covering a large population for severe mental illness prediction and increasing efficiency for data usage. We believe findings from this study can be used to inform further screening tool construction and serve as starting points for further investigations involving larger populations and more data sources.

## Conflict of interest

All Authors declare no Competing Financial or Non-Financial Interests

## Author contributions

D.L. and I.K. initialized the project, developed the original ideas and coordinated the research. D.L. and K.C. conducted the analysis. K.C., P.L. and J.W.S provided medical expertise in mental health. D.L., I.K, J.W.S, P.L., W.Y. and K.H.Y designed the study. W.Y. and K.H.Y provided statistical support. All authors contributed to writing.

Supplementary materials

**Supplementary table 1** Axis I diagnoses and their Phecodes

| Axis I diagnoses | Phecode |
|---|---|
| Anxiety disorder | 300.1 |
| Depression disorder | 296.2 |
| Attention deficit hyperactivity disorder | 313.1 |
| Cannabis misuse/dependence | 316 |
| Oppositional defiant disorder | 312 |
| Conduct disorder | 312 |
| Alcohol abuse/dependence | 317 |
| Post-traumatic stress disorder | 300.9 |
| Specific phobia | 300.13 |
| Social phobia | 300.12 |
| Obsessive-compulsive disorder | 300.3 |
| Panic disorder | 300.12 |
| Opiate abuse/dependence | 316 |
| Eating disorder | 305.2 |
| Stimulant abuse/dependence | 316 |
| Polysubstance abuse/dependence | 316 |
| Agrophobia | 300.12 |
| Adjustment disorder | 304 |
| Dysthymic disorder | 300.4 |
| Cocaine abuse/dependence | 316 |